\documentclass[sigconf]{acmart}

\makeatletter
\def\@ACM@checkaffil{
    \if@ACM@instpresent\else
    \ClassWarningNoLine{\@classname}{No institution present for an affiliation}%
    \fi
    \if@ACM@citypresent\else
    \ClassWarningNoLine{\@classname}{No city present for an affiliation}%
    \fi
    \if@ACM@countrypresent\else
        \ClassWarningNoLine{\@classname}{No country present for an affiliation}%
    \fi
}
\makeatother

\AtBeginDocument{%
  \providecommand\BibTeX{{%
    \normalfont B\kern-0.5em{\scshape i\kern-0.25em b}\kern-0.8em\TeX}}}

\setcopyright{acmcopyright}

\copyrightyear{2023}
\acmYear{2023}
\setcopyright{acmlicensed}\acmConference[CIKM '23]{Proceedings of the 32nd ACM International Conference on Information and Knowledge Management}{October 21--25, 2023}{Birmingham, United Kingdom}
\acmBooktitle{Proceedings of the 32nd ACM International Conference on Information and Knowledge Management (CIKM '23), October 21--25, 2023, Birmingham, United Kingdom}
\acmPrice{15.00}
\acmDOI{10.1145/3583780.3615199}
\acmISBN{979-8-4007-0124-5/23/10}

\usepackage{caption}
\usepackage{subcaption}
\usepackage[ruled,linesnumbered]{algorithm2e}
\usepackage{balance}
\makeatletter
\newcommand{\linebreakand}{%
  \end{@IEEEauthorhalign}
  \hfill\mbox{}\par
  \mbox{}\hfill\begin{@IEEEauthorhalign}
}
\makeatother

\begin{document}

\title{Mitigating Semantic Confusion from Hostile Neighborhood for Graph Active Learning}

\author{Tianmeng Yang}
\authornote{Work performed during the internship at Huawei.}
\email{youngtimmy@pku.edu.cn}
\affiliation{School of Intelligence Science and Technology, Peking University}

\author{Min Zhou}
\authornote{Corresponding authors.}
\email{zhoumin27@huawei.com}
\affiliation{Huawei Cloud}

\author{Yujing Wang}
\authornotemark[2]
\author{Zhengjie Lin}
\email{{yujwang,zhengjielin}@pku.edu.cn}
\affiliation{Peking University}

\author{Lujia Pan}
\email{panlujia@huawei.com}
\affiliation{Huawei Noah's Ark Lab}

\author{Bin Cui}
\email{bin.cui@pku.edu.cn}
\affiliation{Peking University}

\author{Yunhai Tong}
\email{yhtong@pku.edu.cn}
\affiliation{Peking University}

\renewcommand{\shortauthors}{Yang, Tianmeng, et al.}

\begin{abstract}
Graph Active Learning (GAL), which aims to find the most informative nodes in graphs for annotation to maximize the Graph Neural Networks (GNNs) performance, has attracted many research efforts but remains non-trivial challenges. One major challenge is that existing GAL strategies may introduce semantic confusion to the selected training set, particularly when graphs are noisy. Specifically, most existing methods assume all aggregating features to be helpful, ignoring the semantically negative effect between inter-class edges under the message-passing mechanism. In this work, we present Semantic-aware Active learning framework for Graphs (SAG) to mitigate the semantic confusion problem. Pairwise similarities and dissimilarities of nodes with semantic features are introduced to jointly evaluate the node influence. A new prototype-based criterion and query policy are also designed to maintain diversity and class balance of the selected nodes, respectively. Extensive experiments on the public benchmark graphs and a real-world financial dataset demonstrate that SAG significantly improves node classification performances and consistently outperforms previous methods. Moreover, comprehensive analysis and ablation study also verify the effectiveness of the proposed framework. 
\end{abstract}




\maketitle
\section{Introduction}
Graph Neural Networks (GNNs) have achieved great success in representation learning for graphs and been widely applied in multiple applications~\cite{ying2018graph,li2022dual,liu2021intention,zhou2022telegraph,yang2022hrcf,liu2022discovering}. Although most existing GNNs models~\cite{yang2016revisiting,kipf2016semi,velivckovic2017graph,huang2021scaling,zhou2023hyperbolic} are trained in a semi-supervised manner, plenty of high-quality training samples are still needed for annotation to achieve satisfactory performance. However, in real-world scenarios such as financial transactions, data is often prone to be scarce of labels, thus leading to high labeling costs with non-negligible human efforts. Active learning (AL), which attempts to maximize a model’s performance gain while annotating the fewest samples possible, has demonstrated benefits across different domains such as computer vision~\cite{sener2017core-set} and natural language processing~\cite{dor2020active-bert}. 

Previous graph active learning methods~\cite{cai2017active,wu2019active,liulscale,zhang2021alg,zhang2021grain} attempt to design different graph-based criteria for node selection on graphs. For example, AGE~\cite{cai2017active} proposes linearly combining uncertainty, graph centrality, and information density scores. ALG~\cite{zhang2021alg} considers node representativeness and informativeness to maximize the effective reception field of GNNs. The recently proposed Grain~\cite{zhang2021grain} connects active learning on graphs with the social influence maximization problem~\cite{kempe2003maximizing} and achieves state-of-the-art performance, with followed works considering incorrect labels or soft labels~\cite{zhang2021rim,yan2023unreal}. Under the GNN's message-passing mechanism which generates feature of the center node by aggregating its neighbors ~\cite{yang2016revisiting,kipf2016semi,velivckovic2017graph}, Most of these approaches assume all aggregating neighborhood features to be friendly and helpful.

However, real-world graphs are often noisy with hostile connections between neighborhood nodes~\cite{kim2022find,liu2021pick,yang2022graph,li2022bsal}. Strategies adopted in existing methods may introduce semantic confusion to the selected training set. Generally, different classes of nodes have variant attribute features implying different semantics.  
Mixing features from inter-class hostile neighbors would lead to a indistinguishable node representation and thus hinder model training.

To mitigate the semantic confusion, we propose a \textbf{S}emantic-aware \textbf{A}ctive learning framework for \textbf{G}raphs (SAG) that captures the complicated node affinities. Specifically, pairwise similarities and dissimilarities of neighborhood features are first introduced to evaluate the node pair influence jointly. Then, a prototype-based diversity criterion is proposed to cover diverse and difficult instances. Each unlabeled node's semantic influence and diversity are carefully unified to distinguish the most informative node. In addition, we also design a class-balanced query policy to finally select nodes for annotation, ensuring the number of selected nodes in each class is balanced.

we conduct extensive experiments on 3 widely used benchmark graphs and a real-world financial fraud detection dataset. The results show that our method consistently outperforms baselines, with a significant lift over the previous state-of-the-art methods. Analysis and ablation study are also performed to reveal the effectiveness and advantages of the proposed SAG framework. In particular, numerical comparison on financial dataset shows that SAG improves by 3.4\% on the binary F1 score and 3.0\% on the AUC score over the best previous methods, respectively, which indicates our model's strength and potential in industrial applications. 

\section{Preliminary}
\paragraph{GAL Problem Formulation. }
Graph Active Learning (GAL) aims to select the most informative nodes to train a GNN model and thus achieves higher performance. Given a graph $\mathcal{G=(V,E)}$, along with its feature matrix $X$, an oracle to label the query node, and a labeling budget $\mathcal{B}$, the validation set is $\mathcal{V}_{val}$, test set is $\mathcal{V}_{test}$, and remaining unlabelled nodes pool are denoted as $\mathcal{U}$. For each iteration, active learning strategies are applied to select one or a batch of nodes from $\mathcal{U}$, query the labels and add to the train set $\mathcal{L}$. This process will be repeated until nodes in $\mathcal{L}$ reach the budget $\mathcal{B}$. The GNN model will then be trained to converge and evaluate the final classification performance.

\section{The Proposed SAG Framework}

\subsection{Semantic-aware Influence}
\begin{figure}[htbp]
	\centering
        \includegraphics[width=\linewidth]{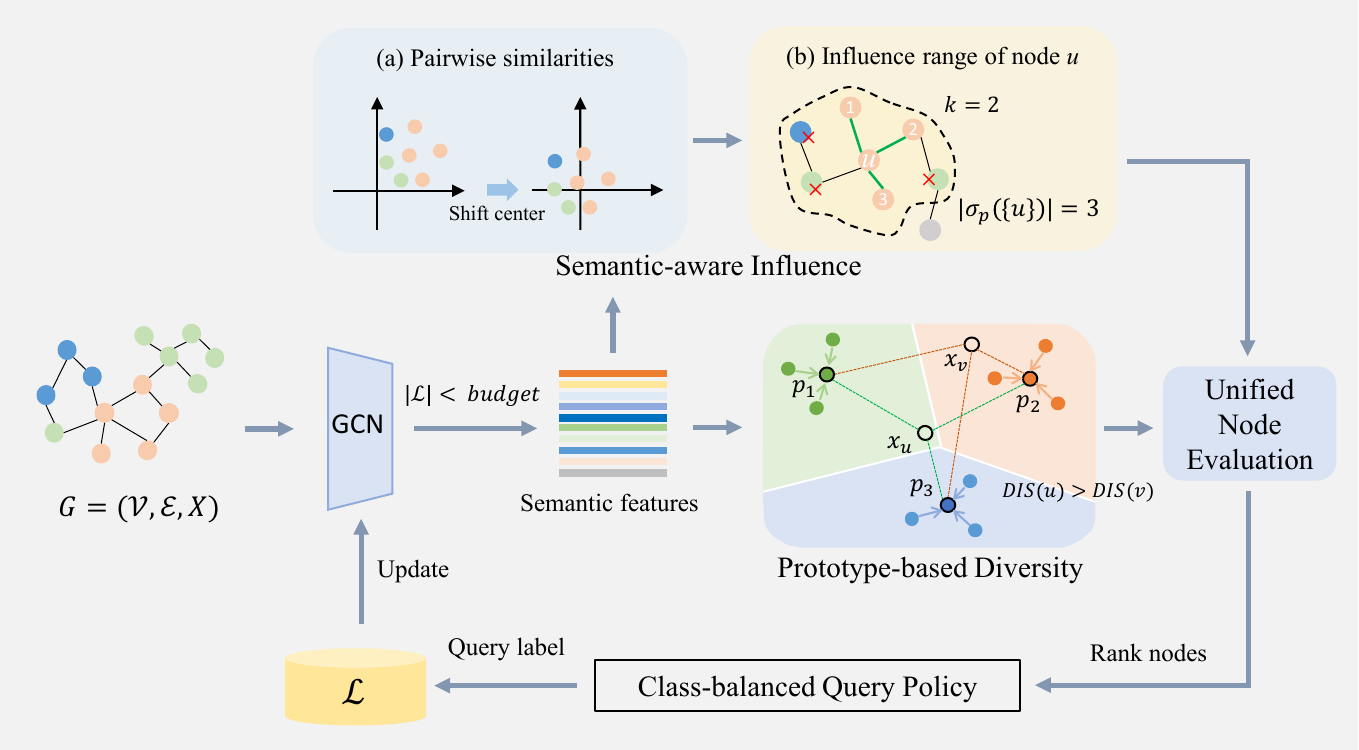}
        \caption{The overall SAG framework.
        }
        \label{fig:archtecture}
\end{figure}

In the neighborhood with inter-class edges, aggregating messages from neighbors of the different classes would mix up features and introduce semantic confusion to node embedding. Our motivation is to select nodes that transfer more clear supervised signals to neighbors instead of confusion. As illustrated in Fig.\ref{fig:archtecture}, SAG first introduce a function $sim(\cdot,\cdot)$ to measure semantic feature similarities between a central node and its neighbors, and distinguish whether the interactions are positive or negative between node pairs. SAG is flexible in applying different $sim(\cdot,\cdot)$ functions to capture node affinities such as attention~\cite{velivckovic2017graph,vaswani2017attention} and Euclidean distance. We also provide an analysis of variant similarity measurements in our experiments in section 4.2. In this work, the cosine similarity is adopted for its better performances. 

Further considering the influence intensity of node interactions, we follow \cite{xu2018representation,wang2020unifying} and leverage the influence function in the statistics field to measure the influence of a specific node on the other nodes. Node influence from node $u$ to $v$ is defined as the L1-norm of the expected Jacobian matrix $\partial x_{v}^{(k)} / \partial x_{u}^{(0)}$ after $k$ layers GCN propagating:
\begin{align}
    I(u,v) = \Vert \mathbb{E}[\partial x_{v}^{(k)} / \partial x_{u}^{(0)}]\Vert_1 .
    \label{Eq:inf}
\end{align}
Both the semantic similarity and influence intensity are contributed to capturing complicated node affinities, and we define the semantic-aware influence of a node as follows:
\begin{align}
    SI(u,v) = sim(u,v) \cdot I(u,v) .
    \label{Eq:SI}
\end{align}

\noindent\textbf{Positive Influence Maximization.} With a threshold $\theta>0$ and a set of selected training nodes $\mathcal{L}$, the positively activated nodes set $\sigma_p(\mathcal{L})$ are defined as :
\begin{align}
    \sigma_p(\mathcal{L}) = \bigcup_{v_l \in \mathcal{L}, v \in N_k(v_l), SI(v_l,v)>\theta} \{v\} ,
\end{align}
where $v_l$ is labeled node in $\mathcal{L}$ and $N_k(v_l)$ is its influence range with a $k$ layers GCN. At each selecting iteration $i$, we hope to find nodes that maximize $\sigma_p(\mathcal{L})$ with more positive influence, thus we can derive the influence increment score of each unlabeled node $u$ by:
\begin{align}
    INF(u) = \left| \sigma_p(\mathcal{L}_i\cup\{u\}) - \sigma_p(\mathcal{L}_i)\right| .
    \label{Eq:sinf}
\end{align}
Larger $INF$ score indicates less semantic confusion as the central node is connected with more friendly nodes possessing similar features.

\subsection{Prototype-based Diversity}
Influence score reflects the capability of activating more nodes when training. To enable the model to benefit from more diverse informative samples, we propose a new prototype-based diversity criterion for node evaluation. Specifically, we maintain a prototype $p_c$ as center of labeled nodes for each class $c$ in each selecting iteration, and encourage the selected training set to explore the most unknown regions that distant from the prototypes. Consequently, we denote the diversity score $DIS(u)$ as the minimum distance between each unlabeled node $u$ and the prototypes.
Nodes far away from the prototypes (i.e., with greater diversity score) indicate that they need to be paid more attention.

\begin{table*}[t]
    \caption{Statistics and properties of the datasets.}
    \label{tab:datasets}
    \centering
    \small
    \scalebox{1.0}{
    \begin{tabular}{lccccccc}
    \toprule
    \textbf{Dataset} & \#Nodes & \#Edges & \#Inter-Class Edges &\# Inter-Class Ratio & \#Features & \#Classes & \#Train/Val/Test\\
    \midrule
    Cora & 2708 & 5429 & 1003 & 18.5\% & 1433 & 7 & 1208/500/1000\\
    Citeseer & 3327 & 4732 & 1204 & 25.4\% & 3703 & 6 & 1827/500/1000\\
    Pubmed & 19717 & 44338 & 8759 & 19.8\% & 500 & 3 & 18217/500/1000\\
    Hpay & 14273 & 80883 & 25988 & 32.1\% & 50 & 2 & 12773/500/1000\\
    \bottomrule
    \end{tabular}
    }
\end{table*}

\begin{table*}[t]
    \caption{Mean metric $\pm$ stdev over ten different runs. The best result on per benchmark is highlighted.}
    \label{tab:experiment}
    \renewcommand\arraystretch{1}
    \centering
    \small
    \resizebox{0.88\textwidth}{23mm}{
    \begin{tabular}{l|c|c|c|c|c|c|c|c}
    \toprule
    \textbf{Dataset} & \multicolumn{2}{c|}{Cora} & \multicolumn{2}{c|}{Citeseer} & \multicolumn{2}{c|}{Pubmed} & \multicolumn{2}{c}{Hpay}\\
    \midrule
    Metrics & Accuracy & Macro-F1 & Accuracy & Macro-F1 & Accuracy & Macro-F1 & Binary F1 & AUC score \\
    \midrule
    Random & 79.96{\tiny$\pm$1.11} & 77.26{\tiny$\pm$2.65} & 70.69{\tiny$\pm$1.27} & 62.70{\tiny$\pm$1.49} & 76.02{\tiny$\pm$4.13} & 74.72{\tiny$\pm$4.73} & 73.78{\tiny$\pm$6.22} & 84.60{\tiny$\pm$3.20} \\
    Degree & 79.41{\tiny$\pm$0.69} & 77.02{\tiny$\pm$0.76} & 63.75{\tiny$\pm$2.24} & 54.81{\tiny$\pm$2.44} & 73.19{\tiny$\pm$3.95} & 68.94{\tiny$\pm$5.38} & 64.12{\tiny$\pm$9.18} & 81.03{\tiny$\pm$5.09} \\
    Entropy & 80.99{\tiny$\pm$1.56} & 79.91{\tiny$\pm$1.53} & 69.63{\tiny$\pm$1.85} & 65.53{\tiny$\pm$1.93} & 76.69{\tiny$\pm$2.28} & 75.80{\tiny$\pm$2.62} & 78.79{\tiny$\pm$2.75} & 85.29{\tiny$\pm$2.42} \\
    AGE & 80.68{\tiny$\pm$1.28} & 79.50{\tiny$\pm$1.00} & 71.03{\tiny$\pm$1.58} & 65.24{\tiny$\pm$2.71} & 78.87{\tiny$\pm$1.68} & 78.04{\tiny$\pm$1.73} & 68.60{\tiny$\pm$5.80} & 80.96{\tiny$\pm$2.94} \\
    Grain & 81.27{\tiny$\pm$0.60} & 78.49{\tiny$\pm$0.76} & 71.17{\tiny$\pm$0.44} & 64.66{\tiny$\pm$0.65} & \underline{80.08{\tiny$\pm$0.55}} & \underline{78.57{\tiny$\pm$0.70}} & 72.76{\tiny$\pm$3.95} & 83.57{\tiny$\pm$1.47}\\
    \midrule
    SAG(ours) & \textbf{82.87{\tiny$\pm$1.34}} & \textbf{81.30{\tiny$\pm$1.55}} & \textbf{72.65\tiny{$\pm$1.11}} & \textbf{66.32{\tiny$\pm$1.26}} & \textbf{80.77{\tiny$\pm$1.34}} & \textbf{79.09{\tiny$\pm$1.60}} & \textbf{81.48{\tiny$\pm$3.15}} & \textbf{87.83{\tiny$\pm$1.17}} \\
    \midrule
    SAG-att & 79.99{\tiny$\pm$2.31} & 78.06{\tiny$\pm$2.63} & 69.74{\tiny$\pm$3.02} & 64.26{\tiny$\pm$3.17} & 77.83{\tiny$\pm$4.20} & 76.76{\tiny$\pm$3.71}  & 77.56{\tiny$\pm$2.68} & \underline{86.52{\tiny$\pm$2.00}} \\
    SAG-dist & \underline{81.79{\tiny$\pm$1.25}} & \underline{80.52{\tiny$\pm$1.22}} & \underline{71.52{\tiny$\pm$1.03}} & \underline{65.64{\tiny$\pm$1.78}} & 78.69{\tiny$\pm$2.21} & 77.65{\tiny$\pm$2.50} & \underline{78.99{\tiny$\pm$2.48}} & 86.47{\tiny$\pm$2.02} \\
    \bottomrule
    \end{tabular}
    }
\end{table*}

\subsection{Unified Node Evaluation}
As influence and diversity mutually drive the evaluation of the unlabeled nodes, we perform score normalization~\cite{zhang2017percent,cai2017active} and combine them in a unified manner. Denoting $\mathcal{P}_{\phi}(u,\mathcal{U})$ as the percentile of nodes in $\mathcal{U}$ which have smaller scores than node $u$ in terms of metric $\phi$, the objective function of SAG to rank and select the node for labeling is defined as:
\begin{align}
    Score(u) = (1-\lambda) \cdot \mathcal{P}_{INF}(u,\mathcal{U}) + \lambda \cdot \mathcal{P}_{DIS}(u,\mathcal{U}) ,
    \label{Eq:overall}
\end{align}
where $\lambda$ is a trade-off coefficient to balance influence and diversity.

\subsection{Class-balanced Query Policy}
Class balance plays an essential role in classification tasks. In a regular semi-supervised setting, the number of labeled nodes in each class is expected to be balanced. 
To this end, after ranking all unlabeled nodes, we utilize a class-balanced query policy to decide whether a node should be labeled. We set an equal budget for each class in advance and acquire the pseudo-label from the classification logits of the node with the highest scores in each iteration. Based on the pseudo-label, once nodes in the class of the current node have reached the budget, this node would be ignored and replaced by the subsequent node whose class budget remains not full yet.

\section{Experiments}

\subsection{Experimental Setup}
\noindent\textbf{Datasets.} We thoroughly evaluate the semi-supervised node classification performance of SAG on the three widely used benchmark graphs - Cora, Citeseer, and Pubmed~\cite{yang2016revisiting,kipf2016semi,velivckovic2017graph}. Moreover, we also collect a real-world financial fraud detection dataset with more noisy edges provided by an online credit payment company, termed as Hpay. The dataset statistics are summarized in Table \ref{tab:datasets}.

\noindent\textbf{Baselines.} We compare SAG with the representative active learning baselines: \textit{Random, Degree, Entropy:} Select nodes randomly,  with maximum degree, and with maximum uncertainty, respectively; \textit{AGE}~\cite{cai2017active}: Combine uncertainty, centrality, and density linearly with time-sensitive parameters; \textit{Grain}~\cite{zhang2021grain}: Select nodes with an diversified influence maximization objective without learning. Here we compare with Grain(ball-D) for which shows higher performances.

\noindent\textbf{Settings.} For a fair comparison, we closely follow the experiment setting in previous work~\cite{cai2017active} . The layers of GCN~\cite{kipf2016semi} backbone are two. Similar to AGE~\cite{cai2017active}, we randomly select 4 nodes as the initially labeled nodes for each class. To enable all the results to be stable and reproducible, the validation set and test set are the same as in~\cite{yang2016revisiting,kipf2016semi,velivckovic2017graph}. We report the average value and standard deviation of 10 runs. For baselines, we use the official settings except for a small change on Grain due to its different test protocol~\footnote{https://github.com/zwt233/Grain/issues/1}. For SAG, the hidden size is 128, the learning rate is 0.05, the dropout rate is 0.5, and the weight decay is 5e-4. Other hyper-parameters are tuned on the validation set.

\begin{figure*}
  \begin{minipage}[t]{\linewidth}
    \begin{subfigure}{0.24\linewidth}
         \centering
         \includegraphics[width=1.0\linewidth]{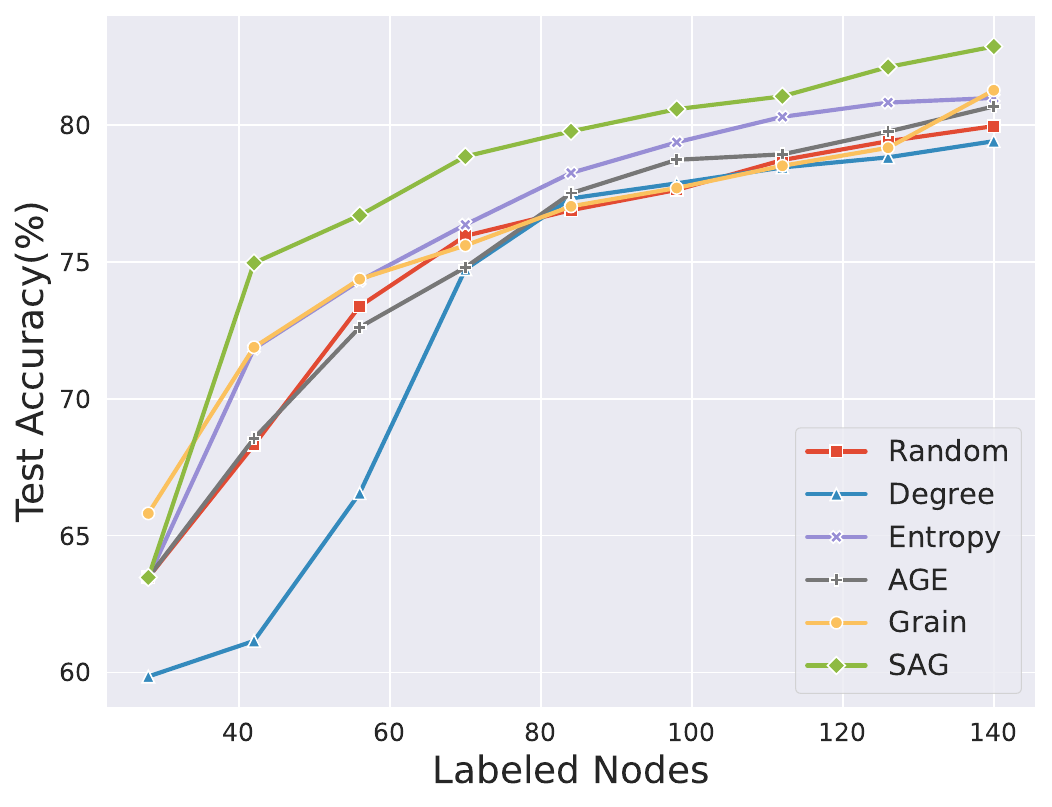}
         \caption{Labeling efficiency}
     \end{subfigure}
     \hfill
    \begin{subfigure}{0.26\linewidth}
         \centering
         \includegraphics[width=1.0\linewidth]{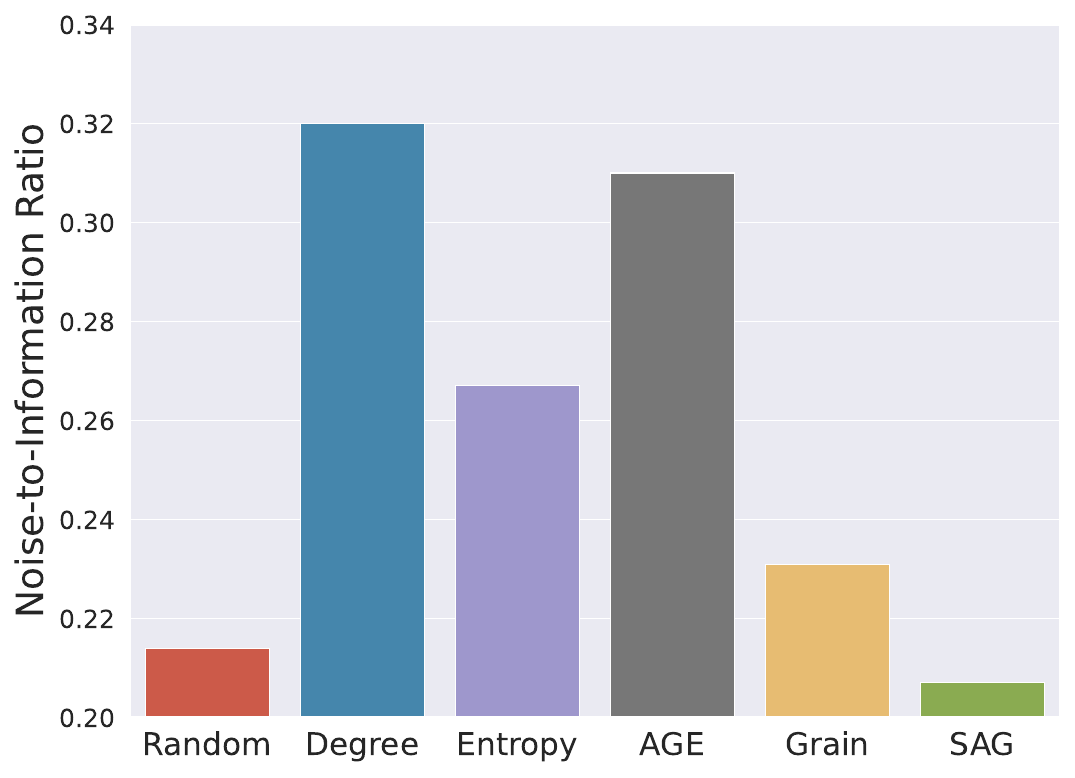}
         \caption{Reducing noise}
     \end{subfigure}
     \hfill
     \begin{subfigure}{0.255\linewidth}
         \centering
         \includegraphics[width=1.0\linewidth]{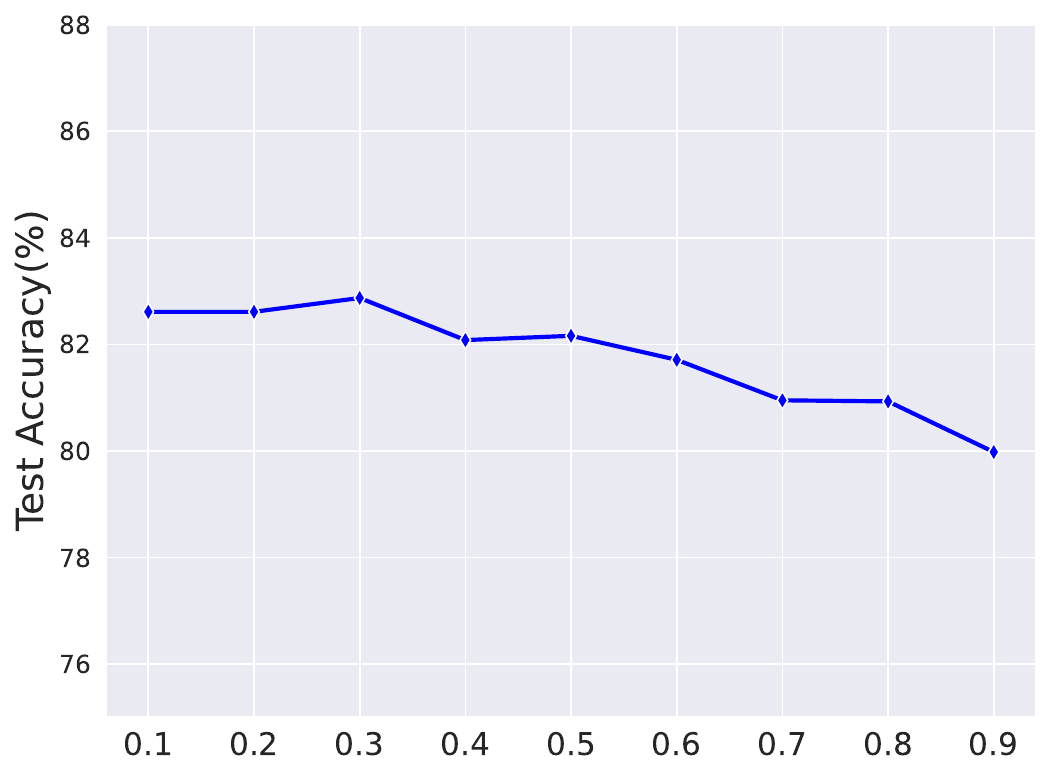}
         \caption{Trade-off coefficient $\lambda$}
     \end{subfigure}
     \hfill
     \begin{subfigure}{0.2\linewidth}
         \centering
         \includegraphics[width=1.0\linewidth]{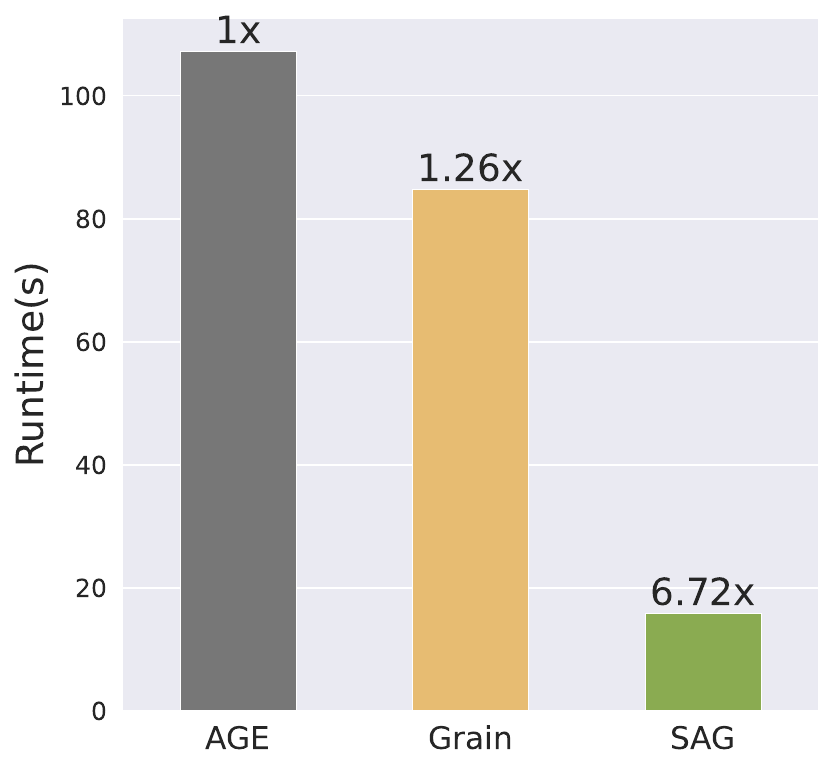}
         \caption{Actual runtime}
     \end{subfigure}
       \caption{Analysis on Cora.}
       \label{fig:analyze}
  \end{minipage}%
\end{figure*}

\subsection{Performance comparison}

As shown in Tabel~\ref{tab:experiment}, most active learning strategies show improvements on the 3 benchmark graphs compared to randomly selecting nodes for training. Our proposed SAG outperforms all baselines by a significant margin, demonstrating its advantages for node classification.

Nevertheless, AGE and Grain fail to select a practical training set for the real-world Hpay dataset, showing inferior performance even compared with random selection. Generally, fraudsters have rich interactions with regular users, thus leading to a severe semantic confusion when aggregating hostile neighborhood features. Besides, fraudsters are usually rare, which makes it important to cope with the heavily imbalanced class distribution. By mitigating the semantic confusion problem and addressing the imbalanced class distribution, SAG improves by 3.4\% on the binary F1 score and 3.0\% on the AUC score over the best baseline results, respectively. This verifies the effectiveness of SAG on noisy graphs and indicates the strength of our model in the field of financial fraud detection.

\textbf{Variant similarity measurements.} The SAG framework is flexible to leverage different similarity measurements capturing node affinities. Here we investigate two typical variants on all datasets for comparison: \textbf{SAG-att} and \textbf{SAG-dist}. We replace the cosine similarity function with attention and euclidean distance, respectively.
We can observe in Table \ref{tab:experiment} that SAG-dist also outperforms most of the baselines, though still lower than SAG. SAG-att shows inferior performance, possibly because the attention model cannot be well joint-optimized in a ranking-based indifferentiable selection process. These results indicate the importance of adequately modeling semantic similarity into active learning.

\subsection{Model Analysis}
We have performed comprehensive analysis to verify the advantages and effectiveness of SAG. Limited by the page, we report main results on Cora as following:

\noindent\textbf{Labeling efficiency.} The first demand of active learning is reducing labeling costs. Denoting $C$ as the number of classes for each dataset, we conduct experiments with different labeling budgets ranging from $4C$ to $20C$ (Note that $4C$ nodes are initially sampled except Grain and Degree). As illustrated in Fig.~\ref{fig:analyze}(a), SAG consistently outperforms the baselines as the number of labeled nodes grows. For example, the competitive baseline Grain achieves an accuracy of 81.27\% with 140 nodes labeled, while SAG only needs 112 nodes to achieve similar results, reducing 20\% human labeling effort.

\noindent\textbf{Reducing noise.} Noises introduced by inter-class edges are usually detrimental to the node representation. Our motivation is that SAG can select nodes with the maximum positive influence and mitigate training confusion. To verify that, we quantitatively define the Noise-to-Information Ratio (NIR) as the proportion of inter-class node pairs in all node pairs that have interactions, and conduct analysis to compare the NIR value of selected nodes among different active learning methods. As Fig.~\ref{fig:analyze}(b) shows, AGE and Grain ignore the negative node affinities and may tend to select nodes with a greater degree, introducing more noise even compared with random selection. SAG explicitly filters the noisy connections, thus being able to find the most informative nodes as well as reducing semantic confusion, achieving the minimum NIR value.

\noindent\textbf{Trade-off coefficient.} We also conduct an analysis of the trade-off coefficient $\lambda$ that unifies the influence and diversity. As shown in Fig.~\ref{fig:analyze}(c), the performance rises first and then starts to drop slowly when $\lambda$ ranges from 0.1 to 0.9. It is probably because the increasing weight of diversity would encourage the AL model to select more uncertain or indistinguishable instances and introduce severe semantic confusion when exceeding a limit.   

\noindent\textbf{Runtime comparison.} Runtime is crucial for active learning on graphs because real-world graphs can become very large. We demonstrate SAG's high efficiency by comparing the end-to-end runtime, including preprocessing with AGE and Grain. As shown in Fig.~\ref{fig:analyze}(d), SAG achieves a speedup of 6.72x over AGE, and 5.33x over Grain, respectively. Note that density score in AGE is computed after clustering, and diversity functions in Grain compute covered balls for each node in every iteration, leading to an $O(N^2)$ time complexity. SAG simplifies the diversity estimation by computing similarities between nodes and prototypes, thus keeping higher efficiency even applied to larger graphs.

\begin{figure} 
    \hfill
    \includegraphics[width=0.98\linewidth]{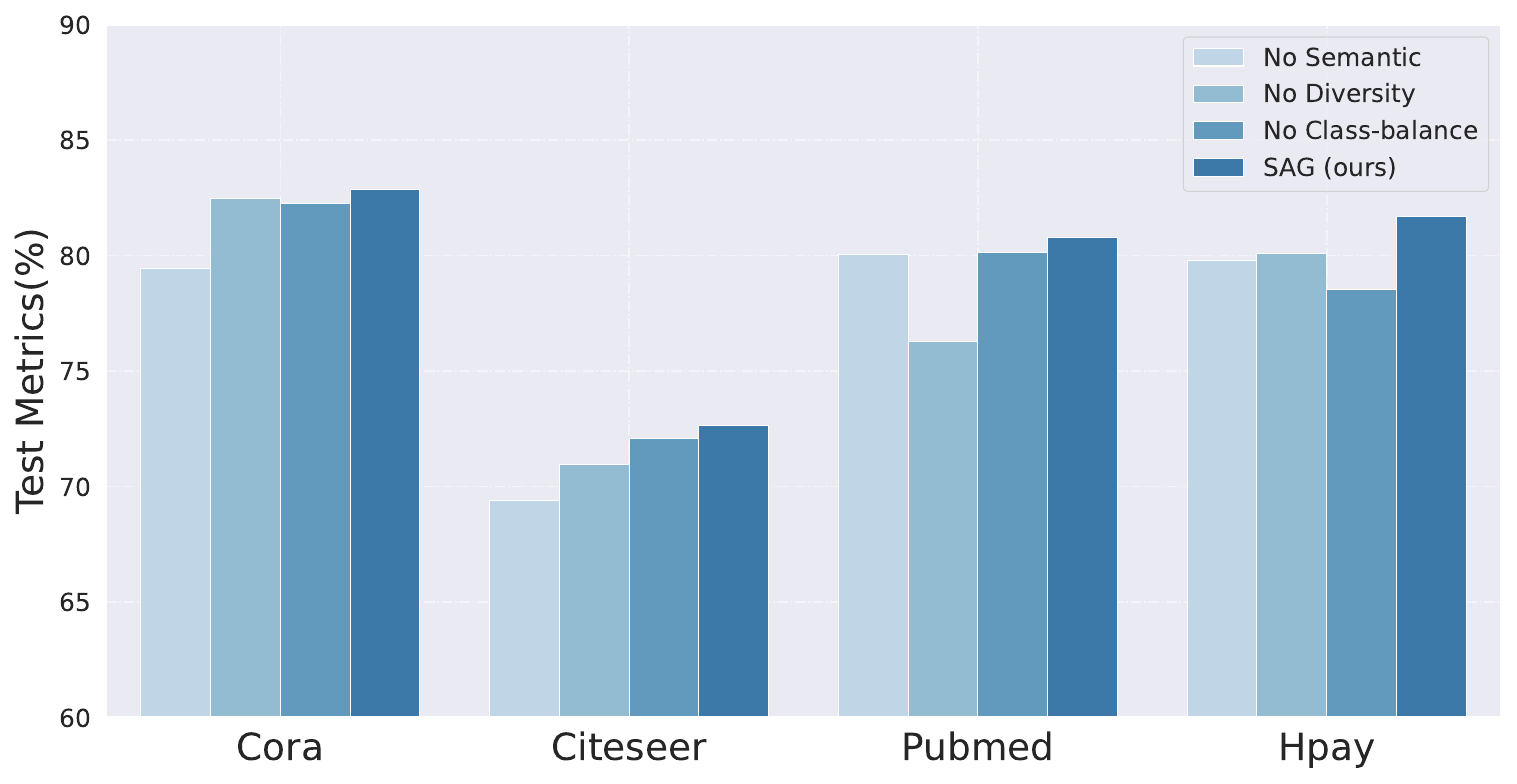}
    \caption{Ablation study.}
    \label{fig:ablation}
\end{figure}

\subsection{Ablation Study}
To verify the effectiveness of each proposed component, we evaluate SAG and its  variants by removing each part: \textbf{(1) No Semantic}: remove the pair similarity when computing node influence; \textbf{(2) No Diversity}: remove the diversity and rank nodes with only the $INF$ score; \textbf{(3) No Class-balance}: remove the class-balanced query policy and select node with the maximum unified score.  

As illustrated in Fig.~\ref{fig:ablation}, removing the pair similarities leads to a significant performance reduction on both Cora and Citeseer, revealing the importance of modeling the semantic relationship when aggregating neighborhood information. In particular, there is an accuracy gap of 3.42\% and 3.24\% on the two graphs, respectively. Besides, as node interaction plays a less critical role on Pubmed~\cite{zhu2020beyond,liu2021non}, the prototype-based diversity shows a more significant effect with an accuracy gap of 4.51\% when removing the diversity module. It is also observed that the class-balanced query policy consistently improves classification performance, especially for the financial fraud detection dataset which is heavily imbalanced. Identifying semantic similarity, meanwhile, is another key to improve performances for Hpay. On all datasets, the full model achieves the best performance, verifying the necessity of our proposed methods.

\section{Conclusion}
In this paper, we focus on mitigating semantic confusion from noisy hostile neighborhood for graph active learning and propose an effective framework termed SAG. Semantic pair similarities are explicitly introduced for node interactions and evaluations. A new prototype-based diversity criterion is also unified with the maximum positive influence to rank unlabeled nodes, followed by a simple yet effective class-balanced query policy. Comprehensive experiments and analysis demonstrate the effectiveness and advantages of SAG, showing its potential for financial industry applications.


\bibliographystyle{ACM-Reference-Format}
\balance
\bibliography{sample-base}




\end{document}